\documentclass[sigconf]{acmart}
\AtBeginDocument{%
  \providecommand\BibTeX{{%
    \normalfont B\kern-0.5em{\scshape i\kern-0.25em b}\kern-0.8em\TeX}}}

\setcopyright{acmcopyright}
\copyrightyear{2023}
\acmYear{2023}
\acmDOI{10.1145/XXXXXXX.XXXXXXX}

\usepackage{graphicx}
\usepackage{subfigure}
\usepackage{multirow}
\usepackage{float}
\usepackage{algorithm}
\usepackage{bbm}
\usepackage{algorithmicx}
\usepackage{algpseudocode}
\usepackage{amsmath}
\begin{document}

\title{BERT4CTR: An Efficient Framework to Combine Pre-trained Language Model with Non-textual Features  for CTR Prediction}

\author{Dong Wang}
\affiliation{%
  \institution{Microsoft STCA}
  \streetaddress{No. 5 Danling Street}
  \city{Beijing}
  \country{China}
}
\email{donwa@microsoft.com}

\author{Kav\'{e} Salamatian}
\affiliation{%
  \institution{University of Savoie}
  \streetaddress{27 rue marcoz, 73000}
  \city{Annecy}
  \country{France}
}
\email{Kave.salamatian@univ-smb.fr}

\author{Yunqing Xia}
\affiliation{%
  \institution{Microsoft STCA}
  \streetaddress{No. 5 Danling Street}
  \city{Beijing}
  \country{China}
}
\email{yxia@microsoft.com}

\author{Weiwei Deng}
\affiliation{%
  \institution{Microsoft STCA}
  \streetaddress{No. 5 Danling Street}
  \city{Beijing}
  \country{China}
}
\email{dedeng@microsoft.com}

\author{Qi Zhang}
\affiliation{%
  \institution{Microsoft STCA}
  \streetaddress{No. 5 Danling Street}
  \city{Beijing}
  \country{China}
}
\email{qizhang@microsoft.com}

\newcommand{\tabincell}[2]{\begin{tabular}{@{}#1@{}}#2\end{tabular}}

\begin{abstract}
Although deep pre-trained language models have shown promising benefit in a large set of industrial scenarios, including Click-Through-Rate (CTR) prediction, how to integrate pre-trained language models that handle only textual signals into a prediction pipeline with non-textual features is challenging.  

Up to now two directions have been explored to integrate multi-modal inputs in fine-tuning of pre-trained language models. One consists of  fusing the outcome of language models and non-textual features through an aggregation layer, resulting into ensemble framework, where the cross-information between textual and non-textual inputs are only learned in the aggregation layer. The second one consists of  splitting non-textual features into fine-grained fragments and transforming the fragments to new tokens combined with textual ones, so that they can be fed directly to  transformer layers in language models. However, this approach increases the complexity of the learning and inference  because of the numerous additional tokens.

To address these limitations, we propose in this work a novel framework BERT4CTR, with the Uni-Attention mechanism that can benefit from the  interactions between non-textual and textual features while maintaining low time-costs in training and inference through a dimensionality reduction. Comprehensive experiments on both public and commercial data  demonstrate that BERT4CTR can outperform significantly the state-of-the-art frameworks to handle multi-modal inputs and be applicable to CTR prediction.


\end{abstract}


\maketitle

\section{Introduction} \label{sec:intro}
Machine learning has frequently to deal with multi-modal inputs that are mixing  numerical, ordinal, categorical, and textual data. This is especially the case for the Click-Through-Rate (CTR) prediction, where one tries to predict the likelihood that a recommended candidate shown after a query entered on a search engine, will be clicked based on not only the semantic relevance between the query and the description of candidate, but also the user's attributes such as user's ID, user's gender, user's category, {\em etc.}, which are non-textual. In recent work, the pre-trained language models like BERT \cite{devlin2018bert} and RoBERTa \cite{liu2019roberta},  which can dig deep for semantic relationship between words in natural language texts, have shown to be beneficial in improving the accuracy of CTR prediction. However, how to integrate pre-trained language models that handle only textual signals into the CTR prediction pipeline with numerous non-textual features is still challenging. Currently these pre-trained language models have been widely used to improve classical CTR prediction, by adding the final score \cite{wang2020gbdt} or intermediate embedding \cite{lu2020twinbert} after fine-tuning with textual signals, {\em e.g.}, $<query, ad>$ textual pairs, as the independent NLP feature into the existing CTR prediction pipelines with non-textual features, leading to cascading workflow. Nonetheless, such frameworks cannot leverage the cross-information between textual and non-textual signals in the fine-tuning of language models. In this paper we confirm that learning such cross-information in fine-tuning phase is beneficial for the improvement on accuracy of CTR prediction and the goal of our work is to design an efficient framework making the information fusion happen at the beginning stage of the fine-tuning process.
It is also noteworthy that although CTR prediction is the main application in this paper, the approach developed here can be extended to a large set of applications that have to deal with multi-modal inputs in pre-trained language models.

Up to now two directions have been explored to integrate multi-modal inputs in fine-tuning of pre-trained language models. In the first approach, called here as "Shallow Interaction", the language model with textual input is treated as a separated and specific network, and the outcome of this network (final output score or [CLS] pooling layer) is fused into the other network dealing with non-textual inputs through an aggregation layer. This approach has been adopted in \cite{chen2019behavior}\cite{muhamed2021ctr}, resulting into ensemble learning framework, and \cite{wang2022learning} has done an in-depth analysis on this approach. In this approach, interaction between textual and non-textual features only happens in the last aggregation layer. As a consequence cross-information between textual and non-textual inputs are not 
dug enough to fine tune the model. In the second class of approach, non-textual features are directly fed as the inputs of transformer layers in the language model, which enables to leverage the non-textual inputs at the beginning stage of model learning. Such an approach is at the core of  VideoBERT \cite{sun2019videobert}, VL-BERT \cite{su2019vl} and NumBERT \cite{zhang2020language}, where  non-textual signals, such as images or numbers, are split into fine-grained fragments ({\em e.g.},  regions-of-interest in images or digits) each of which is transformed as a new token and combined with textual tokens. However, there are always several hundreds of non-textual features in the task of CTR prediction, where the overlong additional
inputs complicate the computations and make the time-costs in learning and inference of model intractable.

Given the limitations of the two approaches in literature, we introduce a simple and light framework, named \textit{BERT4CTR}, to handle multi-modal inputs mixing textual and non-textual features in  pre-trained language models. Our approach is based on a \textit{Uni-Attention} mechanism  integrating the semantic extraction from textual features, with cross-information between textual and non-textual features. We apply a dimensionality reduction operation in order to decrease the time-costs both in learning and inference.  Besides, a two-steps joint-training is introduced to fine-tune the model and further improve the  accuracy of prediction. The proposed approach scales well with the growing of non-textual features, that can be expected to improve the industrial CTR prediction tasks with large number of features.



Through empirical evaluation on both commercial and public data, we show that, compared with these state-of-the-art approaches combining textual and non-textual features in CTR prediction mentioned above, BERT4CTR significantly improves the accuracy of predictions while keeping low latency in training and inference. In particular, our results indicate that by increasing the number of non-textual inputs, the advantages of BERT4CTR are enhanced, {\em e.g.}, on the public data set  with 57 non-textual features, BERT4CTR compared 
with NumBERT shows a significant gain of 0.7\% for the Area Under the ROC Curve (AUC) along with a decrease in training cost of 30\%,  and a decrease in inference cost of 29\%. In the meanwhile, on the  commercial data set with 90 non-textual features, BERT4CTR provides an AUC gain of 0.6\%, along with a decrease in training cost of 64\%, and in inference cost of 52\%.

In section \ref{sec:relatedwork}, we present the related work. The section \ref{sec:bert-ctr}  introduces the design of BERT4CTR. The evaluation is presented in section \ref{sec:experiments}. Finally concluding remarks are provided.
\section{Related Work} \label{sec:relatedwork}
This section presents the related works on multi-modal inputs handling that combine non-textual features with pre-trained language models, and its application to CTR prediction. 

\subsection{Multi-modal Inputs Handling}
The issue of handling inputs that are mixing textual and non-textual input, and integrating semantic insights coming from pre-trained language models like BERT \cite{devlin2018bert} has been already investigated in the literature VideoBERT  \cite{sun2019videobert}, VL-BERT  \cite{su2019vl}, NumBERT  \cite{zhang2020language} and CTR-BERT  \cite{muhamed2021ctr}. The approach followed in these works consists of splitting the non-textual signals into fine-grained fragments each of which is transformed as a new token and combined with textual tokens as the inputs of transformer layers. However, the addition  of tokens representing the non-textual features complicates the language model structure and can make the learning and inference phases too costly for model updating and online serving.

\subsection{Models for CTR Prediction}
CTR prediction is one of the major practical applications of deep learning. Clicks made on advertisements or candidates shown along with search results, or web content presentation, are the main source of revenue for a large set of web actors. In this context, models are always used to select quality advertisements or candidates to present according to the web contents, {\em e.g.}, in sponsored search engines \cite{jansen2008sponsored} or personal recommendation systems \cite{sharma2013survey}, which should achieve both low-latency and high-accuracy. For example, the CTR prediction model in Baidu.com uses a deep neural network, called \textit{Phoenix Nest}, fed with a handcrafted set of features extracted from the user, the query,  and advertisement properties \cite{fan2019mobius}. Google Ads is using the  "Follow The Regularized Leader” (FTRL) model to predict CTR \cite{mcmahan2013ad}, while Google play is using a  Wide \& Deep model described in  \cite{cheng2016wide}. The work \cite{qu2016product} introduces "Product based Neural Network" (PNN) model to capture interactive patterns between features.
This PNN model is extended in \cite{guo2017deepfm} to DeepFM model that emphasizes the interactions between low- and high-order feature. Microsoft Bing.com has adopted a  Neural Network boosted with GBDT ensemble model \cite{ling2017model} for ads CTR prediction. This is the commercial scenario we are considering through this paper. The features used in these CTR prediction models can be grouped into two categories: one is the raw texts from user, query and ad, and the other is the non-textual features including the attributes of users and items such as gender, age, UserId, AdId, {\em etc.} and the outputs generated from sub-models, such as LR model \cite{ling2017model}\cite{mcmahan2013ad}, pre-trained language model \cite{lu2020twinbert}\cite{wang2020gbdt}, {\em etc.}.

\subsection{Application of Pre-trained Language Models in CTR Prediction}
Recent work has shown the abilities of pre-trained language models to extract deep relationship in a sentence pair~\cite{devlin2018bert}\cite{liu2019roberta}\cite{lan2019albert}\cite{sun2019ernie}\cite{liu2019multi}, that are useful for augmenting the semantic features of query and recommendation pair in CTR prediction \cite{lu2020twinbert}\cite{wang2020gbdt}\cite{jiang2020bert2dnn}\cite{guo2020detext}\cite{yu2021cross}. Generally, the pre-trained language models are trained against the real click data, targeting directly the prediction of   click/non-click labels. Thereafter, the score from the final layer  \cite{wang2020gbdt}\cite{jiang2020bert2dnn}, or the embedding  from the intermediate layer \cite{lu2020twinbert}\cite{guo2020detext} of these fine-tuned language models is used as an additional NLP input feature in the CTR prediction model. For example, Microsoft Bing Ads uses the embedding from the hidden layer of TwinBERT model as a semantic feature \cite{lu2020twinbert} while Meituan.com and JD.com use the output score of BERT \cite{wang2020gbdt}\cite{jiang2020bert2dnn}. Besides that cascading framework, some works consider the fusion between the outputs of language models and non-textual features through an aggregation layer, resulting into ensemble learning frameworks (called "Shallow Interaction" in this paper), such as BST \cite{chen2019behavior} and CTR-BERT \cite{muhamed2021ctr}, and \cite{wang2022learning} has done an in-depth analysis on the Shallow Interaction frameworks, where the cross-information between textual and non-textual inputs are not dug enough to fine tune the language models.

\section{Description of BERT4CTR} \label{sec:bert-ctr}
\subsection{Problem Statement}
CTR prediction models are always using multi-modal inputs, mixing $N$ textual features like searching query, titles and URLs of potential ads to show, {\em etc.}, that are denoted as $\mathcal{T} = \{t_1, t_2, ..., t_N\}$, and $M$ non-textual features of different type, {\em e.g.}, dense features such as historical CTR of the query, last click time of the user, {\em etc.}, sparse  features, like ID, category of user, {\em etc.}, denoted as $\mathcal{C} = \{c_1, c_2, ..., c_M\}$.

In traditional usage of pre-trained language models in CTR prediction, where only textual features are used in fine-tuning on click data, the learning process can be formalized as calibrating a network of which the predicted score can approximate the conditional probability of all outcome alternatives, click or non-click for CTR application, given the textual contexts of query and candidates:
\begin{equation} \label{eq:eq1}
  P_{click} = P(click=1|\mathcal{T})
\end{equation}
As stated above, the non-textual features are also crucial in CTR prediction and should not be ignored. When non-textual features are added the conditional probability becomes:
\begin{equation} \label{eq:eq2}
  P_{click} = P(click=1|\mathcal{T}, \mathcal{C})
\end{equation}
The goal in this paper is to design an efficient network structure that can generate scores approximating the distribution $P_{click}$ in Equation \ref{eq:eq2}, while maintaining acceptable training and inference time-costs for industrial application.

\subsection{Model Design}
We then describe the evolution of our proposed network structure, BERT4CTR, by beginning with the NumBERT framework and gradually adding new components to it.
\subsubsection{NumBERT Description} \label{subsubsec:numbert}
\begin{figure}[h]
  \vspace{-1 mm}
  \centering
  \includegraphics[width=\linewidth]{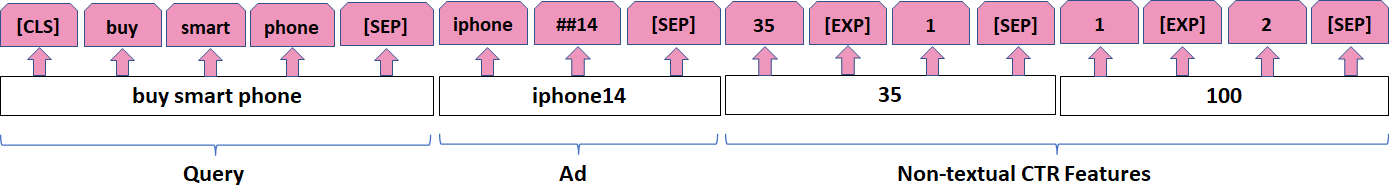}
  \caption{Example of transformed and concatenated non-textual and textual inputs for NumBERT} \label{fig:NumBERTInput}
  \vspace{-2 mm}
\end{figure}

NumBERT \cite{zhang2020language}, is the widely-used systematic approach to integrate textual and numerical features in pre-trained language models. Pre-trained language models like BERT, along with a large class of  neural networks, are using attention layers, that enhance over time some part of the input to enable the training process to concentrate on learning them. 
In each attention layer, a feed-forward network and a residual network are used to control the Vanishing/Exploding gradient issue \cite{he2016deep}.
NumBERT uses a similar structure with several layers of bidirectional self-attention. The core idea in NumBERT is to replace all numerical instances with their scientific notation representations, {\em i.e.}, the number 35 is replaced by "35 [EXP] 1", where [EXP] is a new token that is added to the vocabulary. These transformed non-textual inputs are thereafter considered as normal texts and are fed to the language model. 
For the CTR prediction application, several transformed non-textual inputs might be concatenated using separator token [SEP], to distinguish one numerical feature from other, generating a long string of text which is appended to the end of $<query, ad>$ textual input, and is used for the fine-tuning of language model on click data. 
Figure \ref{fig:NumBERTInput} depicts an example of transformation from original non-textual features to transformed inputs in NumBERT. 

While NumBERT approach enables the language model to understand the numbers in the non-textual signals, the model still misses two crucial elements. First, the contextual relationship between textual features and non-textual ones in CTR prediction are always not obvious. For example, the numerical features such as the historical CTR of user, the ID of user {\em etc.},  are less correlated with $<query, ad>$ texts in semantics. Second issue is related to the fact that the positions of these transformed tokens from non-textual features do not bear semantic meanings as normal texts have, {\em i.e.}, the numerical features in CTR prediction models are always independent of each other.
These two limitations indicate that sharing the same attention weights
and mechanisms for both textual features and the transformed non-textual ones appended are not optimal in fine-tuning, and using simply NumBERT approach to integrate non-textual inputs  cannot improve the performance of learning objectives well, as will be shown later in Section \ref{sec:experiments}. 
\subsubsection{Uni-Attention}
To address these two issues, we have improved the architecture of NumBERT. We are using the same  bidirectional self-attention mechanism as in NumBERT with inputs only from textual tokens. However, for non-textual part, a new type  of attention mechanism is introduced, called {\em Uni-Attention}. It is still a Query-Key-Value (QKV) attention function \cite{bahdanau2014neural}, where the Query is coming only from non-textual tokens, while the Key and Value are coming from textual tokens in the same layer, {\em i.e.}, in the calculation of uni-attention on each token in non-textual part, one input is the matrix projected from the value of that token itself, and the other input is the matrix projected from values of all tokens in textual part. 
In the uni-attention mechanism, the non-textual components have no positional-embedding, which avoids the issue on positional semantics described above. Moreover, this hybrid framework allows the tokens in textual part to dig deep for semantic relationship between each other by the aid of pre-trained attention weights, while grasping the cross-information between textual and non-textual ones in parallel.

\begin{figure}[h]
  \vspace{-1 mm}
  \centering
  \includegraphics[width=\linewidth]{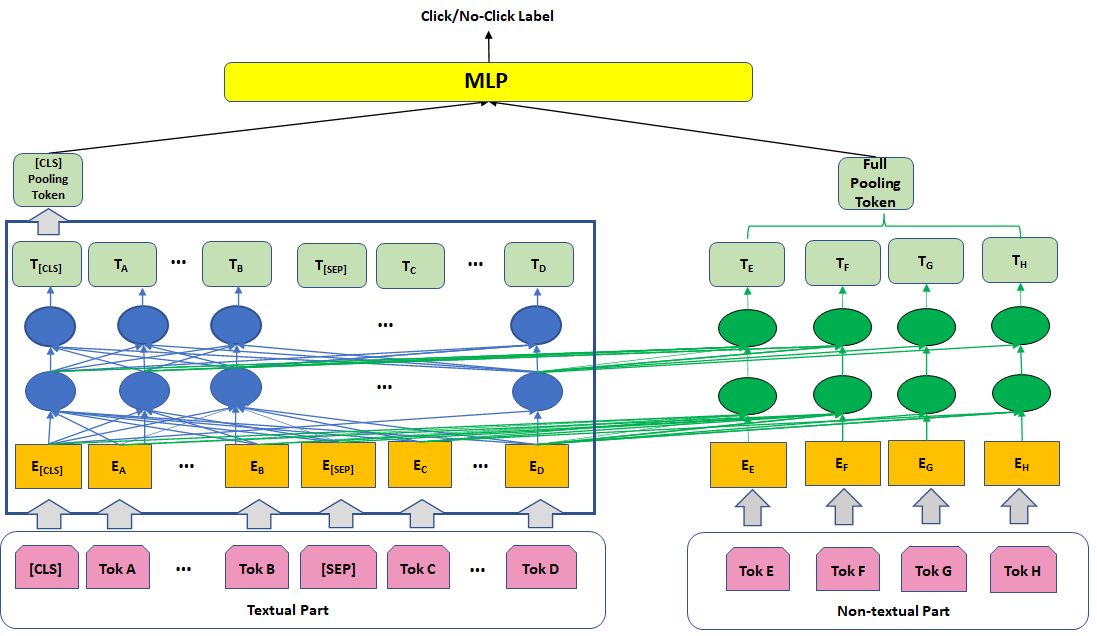}
  \caption{Framework of Uni-Attention} \label{fig:uni-attention}
  \vspace{-2 mm}
\end{figure}

Feed-forward and residual networks are also used on each uni-attention output to control the Vanishing/Exploding gradient issue. We show in Figure \ref{fig:uni-attention} the "Uni-Attention" design. In the last attention layer, all uni-attention outputs from transformed non-textual inputs are gathered as a single hidden layer, which is concatenated with the [CLS] pooling layer from textual part. Thereafter the concatenated layer is fed to a MultiLayer Perception (MLP) that will finally predict the probability of click/non-click.  We will show in Section \ref{sec:experiments} that the proposed design improves strongly the final AUC for both commercial and public data, compared with simple NumBERT.

\subsubsection{Dimensionality Reduction}
The number of non-textual features used in industry for CTR prediction models can be very large, {\em e.g.}, Microsoft Bing Ads uses 90 numerical features transformed each into 4 tokens (accounting for the [EXP] and [SEP] tokens). This large size of inputs impacts negatively  the learning cost and the prediction latency in CTR prediction.

One way to solve this issue is to apply dimensionality reduction on the non-textual features.  Such approaches have already been explored in several previous works like \cite{cheng2016wide}\cite{guo2017deepfm}. Followed
by these works, our approach consists of representing each non-textual feature in $\mathcal{C}$ as a $N$-dimensional point in space. The resulting $N \times |\mathcal{C}|$ space is then mapped to a $K$-dimensional embedding ($K$ $\ll$ $N \times |\mathcal{C}|$) through a fully connected network that is fed, along with the embedding from textual tokens, to the calculation of uni-attentions.

\begin{figure}[h]
  \vspace{-1 mm}
  \centering
  \includegraphics[width=\linewidth]{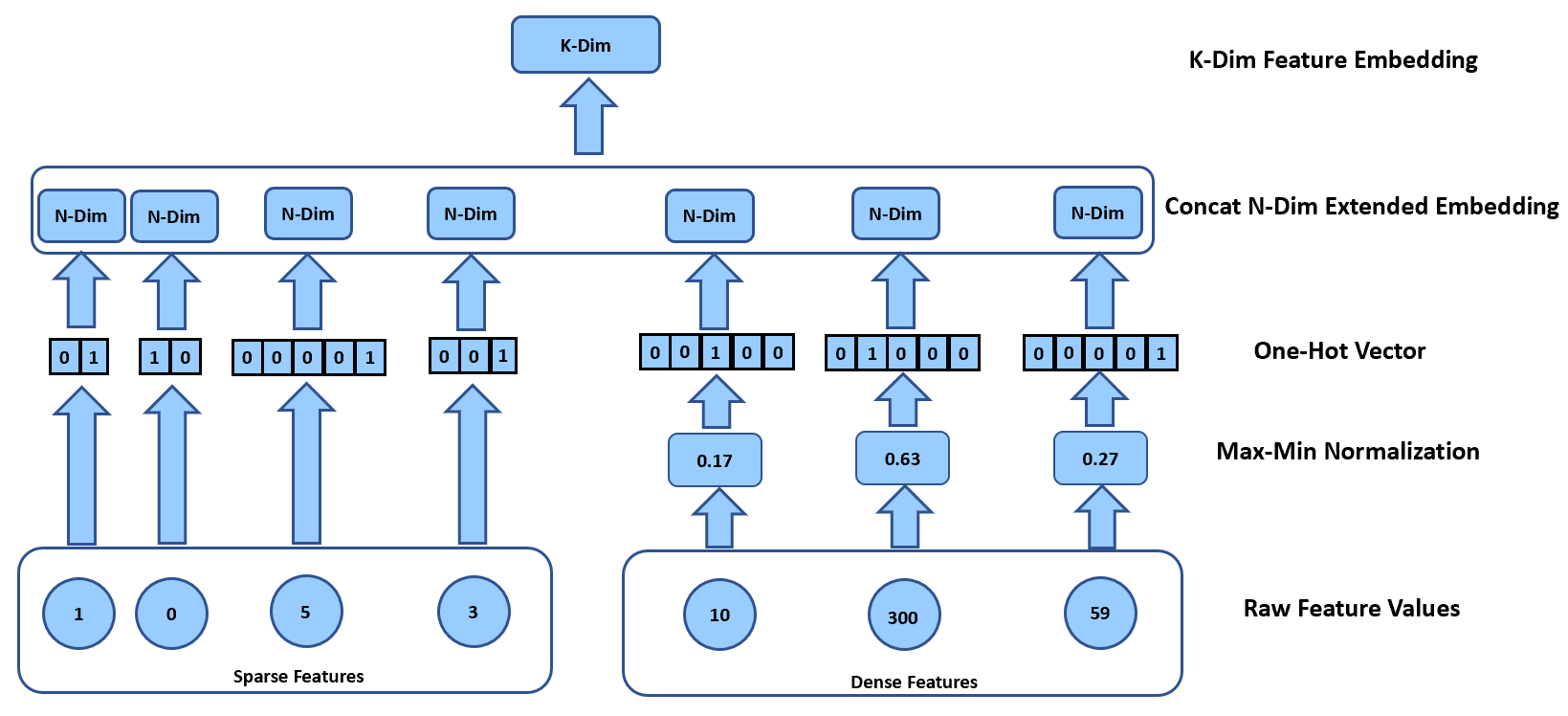}
  \caption{Dimensionality Reduction on embedding of non-texual features in BERT4CTR} \label{fig:feature_embedding}
  \vspace{-2 mm}
\end{figure}

The mapping to the initial $N$-dimensional space is done differently depending if the non-textual features, are dense, {\em e.g.}, the length of the query, the historical value of CTR  {\em etc.}, or sparse,  {\em e.g.},  user's gender, query's category  {\em etc.}.  For sparse features, we use an embedding table that lists the $N$-dimensional embedding corresponding to each given value. Dense features are first normalized using a max-min normalization and thereafter expanded into a 101-dimensional one-hot vectors with 0.01 buckets, that are used as index in an embedding table to find the $N$-dimensional embedding. We show in Figure \ref{fig:feature_embedding} the embedding of non-textual features  used in BERT4CTR.

Similar with NumBERT, the attention alignment score in textual part is calculated as a dot-product form where the dimensions of Query and Key should be the same. However, in the non-textual part after dimensionality reduction, there is not anymore guarantee that the dimension of embedding in non-textual
part is equivalent to the one in textual part.
Considering the flexibility of our model, we introduce  additive attention \cite{bahdanau2014neural} instead of dot-product attention in non-textual part. Additive attention, also known as Bahdanau attention, uses a one-hidden layer feed-forward network to calculate the attention alignment score and
the formula for attention alignment score between Query and Key is as follows:
\begin{equation} \label{eq:additive_attention}
  f_{att}(Q,K) = v_a^Ttanh(W_a[Q;K])
\end{equation}
\noindent where $[Q;K]$ is the concatenation of Query and Key, and  $v_a$ and $W_a$ are learned attention parameters. In \cite{vaswani2017attention} it is shown that  additive attention and dot-product attention are equivalent in computing the attention alignment score between Query and Key, while additive one does not require Query and Key with same embedding dimensions. 

In Section \ref{sec:experiments}, we will show in Table \ref{table:featureembedding} and Table \ref{table:featureembedding_latency} that the dimensionality reduction operation proposed here can hold more than 90\% of the best AUC achieved with uni-attention while substantially reducing the time-costs of training and inference. 

\subsubsection{Two-steps Joint-training} \label{subsubsec:joint-training}
\begin{figure*}[!htbp]
  \vspace{-1 mm}
  \centering
  \includegraphics[width=\linewidth]{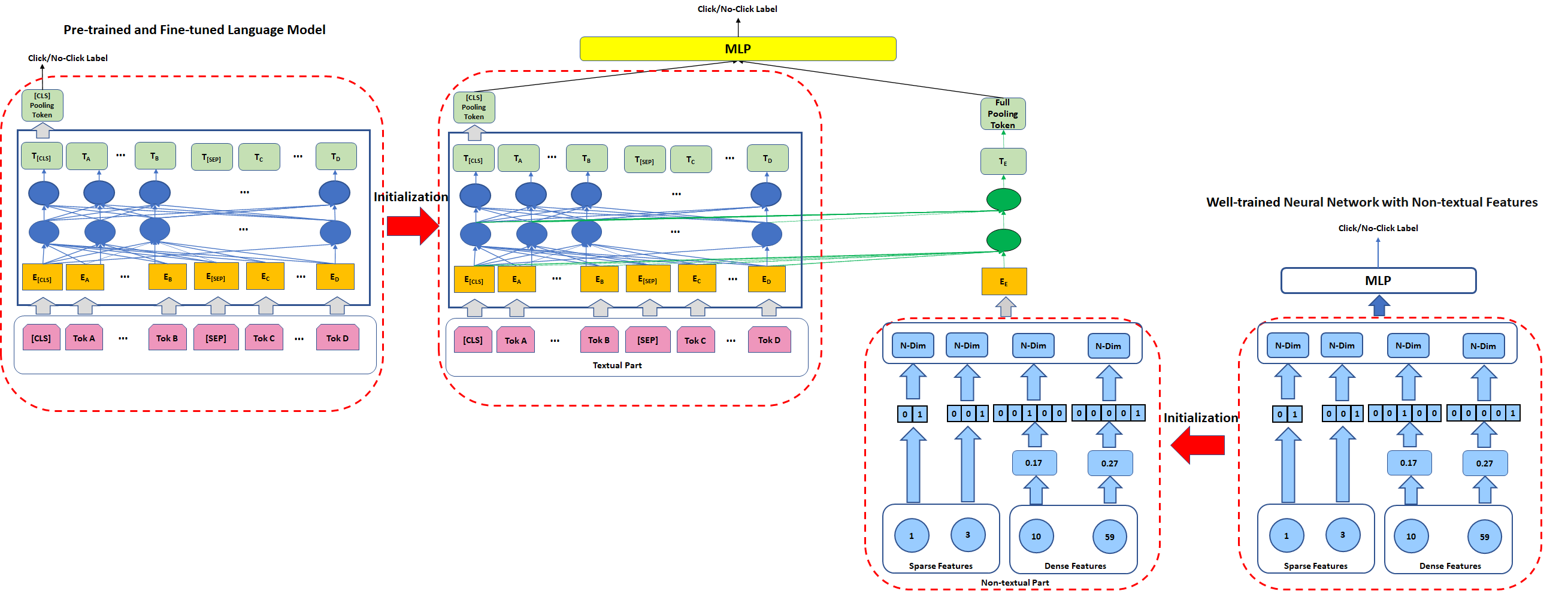}
  \caption{Two-steps Joint-training in BERT4CTR} \label{fig:joint-training}
  \vspace{-2 mm}
\end{figure*}
The calibration of BERT4CTR consists of joint-training with both textual and  non-textual features.  In \cite{wang2022learning} it is shown  that a two-steps training can significantly improve the accuracy of prediction on such joint-training framework and inspired with this, BERT4CTR is also trained in two-steps. In the first step, called \textit{warm-up step}, we pre-train the standard language model with only textual features using a Mask Language Model (MLM) task and then fine-tune this model on the same textual data with click label. The  non-textual part with dimensionality reduction, is also pre-trained using a MultiLayer Perceptron  (MLP) that predicts the click probability using non-textual features alone. This pre-training phase will calibrate the dimensionality reduction parameters.  A cross entropy loss function is used for this learning. The second step of training, called  \textit{joint-training step}, is initialized with the pre-trained textual model, as well as the pre-trained non-textual one, and continues training the whole network of BERT4CTR, mixing textual and non-textual inputs, with a small learning rate.

We demonstrate this two-steps joint-training in Figure \ref{fig:joint-training}.  The results in Section \ref{sec:experiments} will show that the two-steps joint-training provides significant AUC gain on both commercial and public data sets.



\section{Experiments and evaluations} \label{sec:experiments}
In this section, we evaluate BERT4CTR over two  data sets: one is from  Microsoft Bing Ads, called commercial data set, and the other one is from KDD CUP 2012, called public data set. At first, we will describe the experimental settings on the data sets, the pre-trained language models, the baselines, the evaluation metrics and the environments used. We then compare four incrementally complete versions of the proposed framework,  followed by the introductions in Section \ref{sec:bert-ctr}, {\em i.e.}, NumBERT alone, with uni-attention added, with dimensionality reduction for non-textual feature embedding added, and with the two-steps joint-training. This will show the improvements coming from each individual component gradually.  

We will also compare BERT4CTR with three current state-of-the-art frameworks that can handle multi-modal inputs in CTR prediction. These comparisons will provide evidences that BERT4CTR is an efficient framework to combine pre-trained language model with non-textual features for CTR prediction. 

\subsection{Experimental Settings}
\subsubsection{Data Sets}
We use the following two data sets for evaluation. Moreover, to evaluate the robustness of our work, the experiments on different data sets are also based on different pre-trained language models.

\textbf{Microsoft Bing Ads Data Set}: Microsoft Bing Ads is a commercial system used by Microsoft to select the ads presented to users after a requested search. This data set consists of 190 million $<query, ad>$ pairs with click labels, which are randomly sampled from Bing Ads logs obtained in April, 2022. The samples in the first three weeks of April are used as training set, and the remaining as validation set.  Similarly to \cite{wang2022learning}, we use the text of query, and ad title concatenated with ad display URL as two textual features.  In addition, a set of 90 non-textual features are also available in this data set, which can be grouped as: (1) dense features representing continuous numerical values, such as historical value of  CTR per user, number of historical impressions per ad  {\em etc.}; (2) sparse features representing discrete values, such as the user's gender, searching query's category {\em etc.}; (3) ads position, a special feature in CTR prediction. As in \cite{ling2017model}\cite{wang2022learning}, the displayed position of an ad is assumed to be independent of other features, {\em i.e.}, we consider that the displayed position and the quality of an ad are independent on the likelihood of a click.

\textbf{KDD CUP 2012 Data Set}\footnote{https://www.kaggle.com/c/kddcup2012-track2}: We also use in our experiments a public data set coming from KDD CUP 2012 Track 2 \cite{pan2019warm}\cite{wang2013advertisement}. This data set contains 235 million  $<query, ad>$ pairs with click label, sampled from the logs of Tencent search engine Soso.com. This data set contains 57 non-textual features which can also be classified into dense and sparse features, along with the position of ads. 

However, in this data set, there is no time information as in Bing Ads, meaning that it is not possible to split the training and validation data based on time. Thus, we have generated the training and validation set by randomly selecting 1/11 of samples as validation data and the remaining as training data. 

\subsubsection{Pre-trained Language Model Settings}
The textual part of Bing Ads data set is initialized over the RoBERTa-Large model with 24 layers (abbreviated as RoBERTa-24) created  by Facebook \cite{liu2019roberta}. The pre-training of the RoBERTa-24 model is done using the popular Mask Language Model (MLM) task similarly to  \cite{devlin2018bert}\cite{wang2022learning}. 

For the textual part in KDD CUP 2012 data set, a BERT-Base model with 12 layers (abbreviated as BERT-12) \cite{devlin2018bert} is pre-trained  with MLM task and then used as initial model for further experiments.  

To enable reproducibility, we present all the details of the experiments run on the KDD CUP 2012 data, including the data pre-processing steps, hyper-parameter settings and pseudo codes, in the appendix.

\subsubsection{Baseline Setups}
We compare here BERT4CTR with three state-of -the-art frameworks handling pre-trained language model and non-textual features  for CTR prediction.

The first baseline framework, called {\em Cascading Framework}, is a traditional way to introduce pre-trained language models in CTR prediction. It consists of injecting the outcome (final score or intermediate embedding) of language model fine-tuned on the textual inputs alone as a new input feature, along with the non-textual features, for the CTR prediction. Here we first fine-tune one language model (RoBERTa-24 or BERT-12) sufficiently with only $<query, ad>$ textual pairs, and then feed the predicted score of such fine-tuned language model as a new feature into a downstream CTR prediction model. To show the  generality of our work, we choose three different CTR prediction models: (1) Wide \& Deep \cite{cheng2016wide}, introduced by Google, which combines a shallow linear model with a deep neural network; (2) DeepFM \cite{guo2017deepfm}, an improved version of Wide \& Deep, which replaces the linear model with a Factorization-Machine (FM); (3) NN boosted GBDT \cite{ling2017model}, used for Microsoft Bing Ads, that consists of a Neural Network (NN) boosted with a Gradient Boosting Decision Tree (GBDT) ensemble model.

The second baseline is called {\em Shallow Interaction Framework}, which is also widely used in practice \cite{chen2019behavior}\cite{wang2020gbdt}\cite{muhamed2021ctr}. It consists of fusing the non-textual embedding layer and the last layer of pre-trained language model, {\em e.g.}, the [CLS] pooling layer, through an aggregation layer. We use two variants of this approach: the first one, called \textit{Shallow Interaction-1 Layer}, connects the language model and the non-textual embedding layer directly through a MultiLayer Perception (MLP). The second one, called \textit{Shallow Interaction-N Layers}, uses the same number of feed-forward network (FFN) and residual network layers stacked above the non-textual embedding layer, as the ones used in the language model, followed by a MLP. The second variant provides a more fair comparison as the depths of network in textual and non-textual part are the same as BERT4CTR. 

The third baseline is the NumBERT framework \cite{zhang2020language} described in Section \ref{subsubsec:numbert}. 

\subsubsection{Evaluation Metrics}
The Area Under the ROC Curve (AUC) \cite{fawcett2006introduction} and Relative Information Gain (RIG) \cite{moore2001information} are two crucial metrics to evaluate the performance of predictive models, which are also used in our evaluations. Besides the measurements
on the whole validation data (called as \textit{ALL Slice}), we also focus on the infrequent $<query, ad>$ pairs (called \textit{Tail Slice}), which could lead to cold starting problem in CTR prediction. As reported in  \cite{guo2017deepfm}, 0.1\% improvement on AUC or RIG can be seen as significant gain for industrial use. We also use t-test results with $\alpha=0.05$ to compare the performances of different models, {\em i.e.}, a difference between two AUCs (or RIGs) with t-value larger than 3 can be considered as significant and confident \cite{livingston2004student}.  

Besides the AUC and RIG, we also use the average, median, 90th percentile and 95th percentile of the time-costs (milliseconds per sample), both for training and inference, as two additional performance metrics. These two metrics are important for CTR prediction in practice. First, the CTR prediction models have to update frequently to adapt to user's interest drift, {\em e.g.}, the CTR model is refreshed weekly in Microsoft Bing Ads. Therefore, the training time should be less than the refreshing interval. Second, the  time-cost in inference is directly related to the online serving latency and should be set as low as possible. It is noteworthy that for different frameworks, the calculations of time-cost are also different. In terms of cascading framework, the time-cost in training/inference is the summation of time-cost in training/inference of language model and the one in training/inference of downstream CTR prediction model. For both Shallow Interaction and BERT4CTR which need two-steps joint-training, the time-costs in training are calculated as the summation of time taken in warm-up step and the one in joint-training step, while the time-cost in training of NumBERT is only considered as the time taken in pre-training and fine-tuning. The time-costs in inference of all these three no-cascading frameworks are measured as the time taken in single prediction by language models.


\subsubsection{Environments}
All model evaluations are implemented with TensorFlow running on NVIDIA V100 GPUs  with 32GB memory. The maximum lengths of sequence for $<query, ad>$ are set as 64, and the batch sizes are set as 10 on both RoBERTa-24 and BERT-12 model. To avoid the random deviation, each experiment for time-cost is repeated for twenty times to obtain metrics. Without explicit statement, all AUCs and RIGs shown in this section are obtained at the best step during training. 

\subsection{Performance of Components in BERT4CTR}
In this section, we evaluate the improvement on model performance coming from each individual component of BERT4CTR described in Section \ref{sec:bert-ctr}. 




\subsubsection{NumBERT's Performance}  \label{subsubsec:exp_numbert}

\begin{table*}[!htbp]
\begin{footnotesize}
\resizebox{2.1\columnwidth}{!}{
\begin{tabular}{|c|c|cc|cc|cc|cc|cc|cc|cc|cc|cc|}
\hline
\multirow{3}{*}{Dateset}      & \multirow{3}{*}{Slice} & \multicolumn{2}{c|}{Model 1}       & \multicolumn{2}{c|}{Model 2}                  & \multicolumn{2}{c|}{Model 3}       & \multicolumn{2}{c|}{\multirow{2}{*}{$\Delta$ $AUC_{M2-M1}$}} & \multicolumn{2}{c|}{\multirow{2}{*}{$\Delta$ $RIG_{M2-M1}$}} & \multicolumn{2}{c|}{\multirow{2}{*}{$\Delta$ $AUC_{M3-M1}$}} & \multicolumn{2}{c|}{\multirow{2}{*}{$\Delta$ $RIG_{M3-M1}$}} & \multicolumn{2}{c|}{\multirow{2}{*}{$\Delta$ $AUC_{M3-M2}$}} & \multicolumn{2}{c|}{\multirow{2}{*}{$\Delta$ $RIG_{M3-M2}$}} \\ \cline{3-8}
   &                        & \multicolumn{2}{c|}{TextOnly}     & \multicolumn{2}{c|}{\begin{tabular}[c]{@{}c@{}}Shallow Interaction\\ - 1 Layer\end{tabular}} & \multicolumn{2}{c|}{NumBERT}       & \multicolumn{2}{c|}{}                                        & \multicolumn{2}{c|}{}  & \multicolumn{2}{c|}{}    & \multicolumn{2}{c|}{}    & \multicolumn{2}{c|}{}     & \multicolumn{2}{c|}{}      \\ \cline{3-20} 
  &                        & \multicolumn{1}{c|}{AUC}   & RIG   & \multicolumn{1}{c|}{AUC}         & RIG        & \multicolumn{1}{c|}{AUC}   & RIG   & \multicolumn{1}{c|}{Diff}              & T             & \multicolumn{1}{c|}{Diff}              & T              & \multicolumn{1}{c|}{Diff}              & T              & \multicolumn{1}{c|}{Diff}              & T              & \multicolumn{1}{c|}{Diff}              & T              & \multicolumn{1}{c|}{Diff}              & T             \\ \hline
\multirow{2}{*}{Bing Ads}     & ALL                    & \multicolumn{1}{c|}{0.8691} & 0.4987 & \multicolumn{1}{c|}{0.8968}       & 0.5360      & \multicolumn{1}{c|}{0.8961} & 0.5348 & \multicolumn{1}{c|}{0.0277}             & 187.59                & \multicolumn{1}{c|}{0.0373}              & 191.72                 & \multicolumn{1}{c|}{0.0270}             & 185.07                & \multicolumn{1}{c|}{0.0361}              & 188.95                & \multicolumn{1}{c|}{-0.0007}             & 2.97                 & \multicolumn{1}{c|}{-0.0012}              & 3.73                 \\ \cline{2-20} 
  & Tail                   & \multicolumn{1}{c|}{0.7703} & 0.4382 & \multicolumn{1}{c|}{0.8084}       & 0.4726      & \multicolumn{1}{c|}{0.8078} & 0.4719 & \multicolumn{1}{c|}{0.0381}             & 181.50                & \multicolumn{1}{c|}{0.0344}              & 183.06                & \multicolumn{1}{c|}{0.0375}             & 178.81                & \multicolumn{1}{c|}{0.0337}              & 180.34                & \multicolumn{1}{c|}{-0.0006}             & 2.39                 & \multicolumn{1}{c|}{-0.0007}                 & 2.96                 \\ \hline
\multirow{2}{*}{KDD CUP} & ALL                    & \multicolumn{1}{c|}{0.7591} & 0.3917 & \multicolumn{1}{c|}{0.8286}       & 0.4842      & \multicolumn{1}{c|}{0.8273} & 0.4827 & \multicolumn{1}{c|}{0.0695}             & 135.68                & \multicolumn{1}{c|}{0.0925}              & 191.72                & \multicolumn{1}{c|}{0.0682}             & 132.17                & \multicolumn{1}{c|}{0.0910}              & 187.66                & \multicolumn{1}{c|}{-0.0013}             & 5.11                 & \multicolumn{1}{c|}{-0.0015}              & 5.62                 \\ \cline{2-20} 
                              & Tail                   & \multicolumn{1}{c|}{0.6757} & 0.2768 & \multicolumn{1}{c|}{0.7537}       & 0.3739      & \multicolumn{1}{c|}{0.7521} & 0.3724 & \multicolumn{1}{c|}{0.0780}             & 142.24                & \multicolumn{1}{c|}{0.0971}              & 185.87                & \multicolumn{1}{c|}{0.0764}             & 136.09                & \multicolumn{1}{c|}{0.0956}              & 182.23                & \multicolumn{1}{c|}{-0.0016}             & 5.15                 & \multicolumn{1}{c|}{-0.0015}              & 4.74                 \\ \hline
\end{tabular}
}
\end{footnotesize}
\caption{AUC and RIG performance of NumBERT on two data sets} \label{table:numbert}
\vspace{-5 mm}
\end{table*}
We present in Table \ref{table:numbert} the performance of NumBERT for CTR prediction over the two data sets used in this paper. In this case two baseline models can be used for comparison. The first one, called \textit{TextOnly}, uses the pre-trained language model fine-tuned with only $<query, ad>$ textual input and without any non-textual features. The second one is the Shallow Interaction-1 Layer described above. We show in the table along with absolute value of AUC and RIG for each model,  the difference of metrics between two models with t-values, {\em e.g.}, $\Delta$ $AUC_{M3-M1}$, the AUC difference between Model 3 (NumBERT) and Model 1 (TextOnly).

One can observe, from Table \ref{table:numbert}, that NumBERT has been able to benefit from non-textual features. It brings, when compared with the model without non-textual features, 2.7\% AUC improvement over Bing Ads data and 6.8\% AUC improvement on KDD CUP 2012 data. However, compared with the Shallow Interaction model, NumBERT does not provide benefits on both AUC and RIG, and shows worse performance. This means that even if NumBERT allows textual and non-textual features to interact through complex bidirectional self-attention with multi-layers, it is not efficient in learning the cross-information between  multi-modal signals.
\subsubsection{Uni-Attention's Performance}  \label{subsubsec:exp_uniattention}
We follow up with evaluating the improvements coming from uni-attention architecture. We show in Table \ref{table:uniattention} the performance achieved by NumBERT compared with \textit{NumBERT + Uni-Attention}, {\em i.e.}, transformed non-textual features (as depicted in Figure \ref{fig:NumBERTInput}) are fed to uni-attention architecture, as shown in Figure \ref{fig:uni-attention}.

\begin{table}[h]
\begin{footnotesize}
\resizebox{\columnwidth}{!}{
\begin{tabular}{|c|c|cc|cc|cc|cc|}
\hline
\multirow{3}{*}{Dataset}  & \multirow{3}{*}{Slice} & \multicolumn{2}{c|}{Model 1}       & \multicolumn{2}{c|}{Model 2}                                                               & \multicolumn{2}{c|}{\multirow{2}{*}{$\Delta$ $AUC_{M2-M1}$}} & \multicolumn{2}{c|}{\multirow{2}{*}{$\Delta$ $RIG_{M2-M1}$}} \\ \cline{3-6}
                          &                        & \multicolumn{2}{c|}{NumBERT}       & \multicolumn{2}{c|}{\begin{tabular}[c]{@{}c@{}}NumBERT \\ + Uni-Attention\end{tabular}} & \multicolumn{2}{c|}{}                                        & \multicolumn{2}{c|}{}                                        \\ \cline{3-10} 
                          &                        & \multicolumn{1}{c|}{AUC}   & RIG   & \multicolumn{1}{c|}{AUC}                               & RIG                               & \multicolumn{1}{c|}{Diff}                & T                 & \multicolumn{1}{c|}{Diff}                & T                 \\ \hline
\multirow{2}{*}{Bing Ads} & ALL                    & \multicolumn{1}{c|}{0.8961} & 0.5348 & \multicolumn{1}{c|}{0.8988}                             & 0.5397                             & \multicolumn{1}{c|}{0.0027}               & 70.06              & \multicolumn{1}{c|}{0.0049}               & 73.91              \\ \cline{2-10} 
                          & Tail                   & \multicolumn{1}{c|}{0.8078} & 0.4719 & \multicolumn{1}{c|}{0.8111}                             & 0.4772                             & \multicolumn{1}{c|}{0.0033}               & 76.08              & \multicolumn{1}{c|}{0.0053}               & 75.08              \\ \hline
\multirow{2}{*}{KDD CUP}  & ALL                    & \multicolumn{1}{c|}{0.8273} & 0.4827 & \multicolumn{1}{c|}{0.8311}                             & 0.4875                             & \multicolumn{1}{c|}{0.0038}               & 82.13              & \multicolumn{1}{c|}{0.0048}               & 80.70               \\ \cline{2-10} 
                          & Tail                   & \multicolumn{1}{c|}{0.7521} & 0.3724 & \multicolumn{1}{c|}{0.7569}                             & 0.3780                             & \multicolumn{1}{c|}{0.0048}               & 94.14              & \multicolumn{1}{c|}{0.0056}               & 85.11              \\ \hline
\end{tabular}
}
\end{footnotesize}
\caption{AUC and RIG performance of Uni-Attention on two data sets} \label{table:uniattention}
\vspace{-5 mm}
\end{table}
Table \ref{table:uniattention} shows that the uni-attention architecture can bring significant AUC and RIG gains, compared with the NumBERT model without uni-attention, over both data sets. For example, the uni-attention architecture can bring additional 0.3\% AUC gain and 0.5\% RIG gain on Tail Slice of Bing Ads data. These gains are even more obvious for KDD CUP 2012 data, where the AUC gain is 0.5\% and the RIG improves by 0.6\% over Tail Slice. All these changes are statistically significant with t-values larger than 70.
\subsubsection{Dimensionality Reduction's Performance}  \label{subsubsec:exp_featureembedding}
Here, we evaluate the impact of dimensionality reduction of non-textual features, shown in Figure \ref{fig:feature_embedding}, that is made mandatory because of the large number of non-textual inputs in industrial CTR prediction models.

The performances of dimensionality reduction on the two data sets are shown in Table \ref{table:featureembedding}, where \textit{NumBERT + Uni-Attention + Dimensionality Reduction} is the NumBERT model with uni-attention framework, as in Figure \ref{fig:uni-attention}, that is completed with a dimensionality reduction operation in non-textual part,  as shown in Figure \ref{fig:feature_embedding}. Table \ref{table:featureembedding} reports that AUC and RIG for both alternative models are close on the two data sets. Besides, no one of the performance differences is statistically significant, {\em i.e.}, the performance equality hypothesis cannot be refuted. 

Besides the accuracy of prediction, the time-costs in training and inference of these two models are also evaluated in Table \ref{table:featureembedding_latency}. One can observe, that dimensionality reduction reduces strongly the time-cost, up to 45\% of training cost and 24\% of inference cost on KDD CUP 2012 data, with  57 non-textual features, and up to 68\% in training and 43\% in inference on Bing Ads data with 90 non-textual features. This means that dimensionality reduction does not entail a significant performance reduction while reducing obviously the time-costs.


\begin{table}[h]
\begin{footnotesize}
\resizebox{\columnwidth}{!}{
\begin{tabular}{|c|c|cc|cc|cc|cc|}
\hline
\multirow{3}{*}{Dataset}  & \multirow{3}{*}{Slice} & \multicolumn{2}{c|}{Model 1}                                                           & \multicolumn{2}{c|}{Model 2}                                                                                         & \multicolumn{2}{c|}{\multirow{2}{*}{$\Delta$ $AUC_{M2-M1}$}} & \multicolumn{2}{c|}{\multirow{2}{*}{$\Delta$ $RIG_{M2-M1}$}} \\ \cline{3-6}
                          &                        & \multicolumn{2}{c|}{\begin{tabular}[c]{@{}c@{}}NumBERT\\ + Uni-Attention\end{tabular}} & \multicolumn{2}{c|}{\begin{tabular}[c]{@{}c@{}}NumBERT \\ + Uni-Attention\\ + Dimensionality Reduction\end{tabular}} & \multicolumn{2}{c|}{}                                        & \multicolumn{2}{c|}{}                                        \\ \cline{3-10} 
                          &                        & \multicolumn{1}{c|}{AUC}                             & RIG                             & \multicolumn{1}{c|}{AUC}                                            & RIG                                            & \multicolumn{1}{c|}{Diff}                 & T                & \multicolumn{1}{c|}{Diff}                 & T                \\ \hline
\multirow{2}{*}{Bing Ads} & ALL                    & \multicolumn{1}{c|}{0.8988}                           & 0.5397                           & \multicolumn{1}{c|}{0.8980}                                          & 0.5393                                          & \multicolumn{1}{c|}{-0.0008}                    & 2.48              & \multicolumn{1}{c|}{-0.0004}               & 1.97              \\ \cline{2-10} 
                          & Tail                   & \multicolumn{1}{c|}{0.8111}                           & 0.4772                           & \multicolumn{1}{c|}{0.8104}                                          & 0.4766                                          & \multicolumn{1}{c|}{-0.0007}               & 2.77              & \multicolumn{1}{c|}{-0.0006}               & 2.46              \\ \hline
\multirow{2}{*}{KDD CUP}  & ALL                    & \multicolumn{1}{c|}{0.8311}                           & 0.4875                           & \multicolumn{1}{c|}{0.8306}                                          & 0.4869                                          & \multicolumn{1}{c|}{-0.0005}                    & 1.86              & \multicolumn{1}{c|}{-0.0006}               & 2.71              \\ \cline{2-10} 
                          & Tail                   & \multicolumn{1}{c|}{0.7569}                           & 0.3780                           & \multicolumn{1}{c|}{0.7563}                                          & 0.3774                                          & \multicolumn{1}{c|}{-0.0006}               & 2.15              & \multicolumn{1}{c|}{-0.0006}               & 1.93              \\ \hline
\end{tabular}
}
\end{footnotesize}
\caption{AUC and RIG performance of Dimensionality Reduction on two data sets} \label{table:featureembedding}
\vspace{-5 mm}
\end{table}

\begin{table*}[!htbp]
\subtable[Training Cost]{
\resizebox{2\columnwidth}{!}{
\begin{tabular}{|c|cc|cc|cc|cc|}
\hline
\multirow{2}{*}{Model}  & \multicolumn{2}{c|}{Average}  & \multicolumn{2}{c|}{Median}   & \multicolumn{2}{c|}{90th percentile} & \multicolumn{2}{c|}{95th percentile}    \\ \cline{2-9} & \multicolumn{1}{c|}{Bing Ads} & KDD CUP & \multicolumn{1}{c|}{Bing Ads} & KDD CUP & \multicolumn{1}{c|}{Bing Ads} & KDD CUP & \multicolumn{1}{c|}{Bing Ads} & KDD CUP \\ \hline
\begin{tabular}[c]{@{}c@{}}NumBERT + Uni-Attention\end{tabular}                             & \multicolumn{1}{c|}{54.05}     & 11.87    & \multicolumn{1}{c|}{53.76}     & 11.78    & \multicolumn{1}{c|}{54.64}     & 11.92    & \multicolumn{1}{c|}{54.95}    & 11.95     \\ \hline
\begin{tabular}[c]{@{}c@{}}NumBERT + Uni-Attention + Dimensionality Reduction\end{tabular} & \multicolumn{1}{c|}{17.32}      & 6.54     & \multicolumn{1}{c|}{17.16}      & 6.49     & \multicolumn{1}{c|}{17.59}      & 6.61     & \multicolumn{1}{c|}{17.97}     & 6.72 \\ \hline
\end{tabular}
}
}
\subtable[Inference Cost]{
\resizebox{2\columnwidth}{!}{
\begin{tabular}{|c|cc|cc|cc|cc|}
\hline
\multirow{2}{*}{Model}  & \multicolumn{2}{c|}{Average}  & \multicolumn{2}{c|}{Median}   & \multicolumn{2}{c|}{90th percentile} & \multicolumn{2}{c|}{95th percentile}   \\ \cline{2-9} & \multicolumn{1}{c|}{Bing Ads} & KDD CUP & \multicolumn{1}{c|}{Bing Ads} & KDD CUP & \multicolumn{1}{c|}{Bing Ads} & KDD CUP & \multicolumn{1}{c|}{Bing Ads} & KDD CUP  \\ \hline
\begin{tabular}[c]{@{}c@{}}NumBERT + Uni-Attention\end{tabular}                             & \multicolumn{1}{c|}{12.34}     & 4.43    & \multicolumn{1}{c|}{12.29}     & 4.37    & \multicolumn{1}{c|}{12.50}     & 4.49    & \multicolumn{1}{c|}{12.69}    & 4.58     \\ \hline
\begin{tabular}[c]{@{}c@{}}NumBERT + Uni-Attention + Dimensionality Reduction\end{tabular} & \multicolumn{1}{c|}{7.05}      & 3.36     & \multicolumn{1}{c|}{7.03}      & 3.32     & \multicolumn{1}{c|}{7.19}      & 3.42     & \multicolumn{1}{c|}{7.31}     & 3.51      \\ \hline
\end{tabular}
}
}
\caption{Time-cost performance (ms/sample) of Dimensionality Reduction on two data sets} \label{table:featureembedding_latency}
\vspace{-5 mm}
\end{table*}


\subsubsection{Two-steps Joint-training's Performance}  \label{subsubsec:exp_jointtraining}
The last component to be evaluated is the two-steps joint-training described in Section \ref{subsubsec:joint-training}. For this purpose, we compare three initialization approaches for the textual and non-textual parts: (1) Pre-trained but not fine-tuned language model for textual part + Random weights in non-textual part (abbreviated as \textit{No Fine-tuned + Randomly Initialized} in Table \ref{table:jointtraining}); (2) Fine-tuned weights in textual part + Random weights in non-textual part (abbreviated as \textit{Fine-tuned + Randomly initialized} in Table \ref{table:jointtraining}), where the weights in the textual part are initialized using the language model pre-trained and fine-tuned on our $<query, ads>$ textual pairs, and random initial weights are used in non-textual part; (3) Two-steps joint-training where both  weights in textual part and non-textual part are initialized with the weights trained in advance as described in Section \ref{subsubsec:joint-training}. This last setting is the one used for the BERT4CTR model introduced in this paper.

\begin{table*}[!htbp]
\resizebox{2\columnwidth}{!}{
\begin{tabular}{|c|c|cc|cc|cc|cc|cc|cc|cc|}
\hline
\multirow{3}{*}{Dataset} & \multirow{3}{*}{Slice} & \multicolumn{2}{c|}{Model 1} & \multicolumn{2}{c|}{Model 2} & \multicolumn{2}{c|}{Model 3} & \multicolumn{2}{c|}{\multirow{2}{*}{$\Delta$ $AUC_{M3-M1}$}} & \multicolumn{2}{c|}{\multirow{2}{*}{$\Delta$ $RIG_{M3-M1}$}} & \multicolumn{2}{c|}{\multirow{2}{*}{$\Delta$ $AUC_{M3-M2}$}} & \multicolumn{2}{c|}{\multirow{2}{*}{$\Delta$ $RIG_{M3-M2}$}} \\ \cline{3-8}
 &  & \multicolumn{2}{c|}{\begin{tabular}[c]{@{}c@{}}No Fine-tuned \\ + Randomly Initialized\end{tabular}} & \multicolumn{2}{c|}{\begin{tabular}[c]{@{}c@{}}Fine-tuned\\ + Randomly Initialized\end{tabular}} & \multicolumn{2}{c|}{\begin{tabular}[c]{@{}c@{}}Two-steps \\ Joint-training \\ (BERT4CTR) \end{tabular}} & \multicolumn{2}{c|}{} & \multicolumn{2}{c|}{} & \multicolumn{2}{c|}{} & \multicolumn{2}{c|}{} \\ \cline{3-16} 
 &  & \multicolumn{1}{c|}{AUC} & RIG & \multicolumn{1}{c|}{AUC} & RIG & \multicolumn{1}{c|}{AUC} & RIG & \multicolumn{1}{c|}{Diff} & T & \multicolumn{1}{c|}{Diff} & T & \multicolumn{1}{c|}{Diff} & T & \multicolumn{1}{c|}{Diff} & T \\ \hline
\multirow{2}{*}{Bing Ads} & ALL & \multicolumn{1}{c|}{0.8980} & 0.5393 & \multicolumn{1}{c|}{0.8985} & 0.5396 & \multicolumn{1}{c|}{0.9014} & 0.5413 & \multicolumn{1}{c|}{0.0034} & 31.37 & \multicolumn{1}{c|}{0.0020} & 32.13 & \multicolumn{1}{c|}{0.0029} & 29.71 & \multicolumn{1}{c|}{0.0017} & 28.74 \\ \cline{2-16} 
 & Tail & \multicolumn{1}{c|}{0.8104} & 0.4766 & \multicolumn{1}{c|}{0.8106} & 0.4769 & \multicolumn{1}{c|}{0.8136} & 0.4801 & \multicolumn{1}{c|}{0.0032} & 42.53 & \multicolumn{1}{c|}{0.0035} & 41.86 & \multicolumn{1}{c|}{0.0030} & 32.41 & \multicolumn{1}{c|}{0.0032} & 39.42 \\ \hline
\multirow{2}{*}{KDD CUP} & ALL & \multicolumn{1}{c|}{0.8306} & 0.4869 & \multicolumn{1}{c|}{0.8315} & 0.4881 & \multicolumn{1}{c|}{0.8347} & 0.4903 & \multicolumn{1}{c|}{0.0041} & 50.62 & \multicolumn{1}{c|}{0.0034} & 41.48 & \multicolumn{1}{c|}{0.0032} & 32.25 & \multicolumn{1}{c|}{0.0022} & 29.59 \\ \cline{2-16} 
 & Tail & \multicolumn{1}{c|}{0.7563} & 0.3774 & \multicolumn{1}{c|}{0.7571} & 0.3783 & \multicolumn{1}{c|}{0.7618} & 0.3821 & \multicolumn{1}{c|}{0.0055} & 52.14 & \multicolumn{1}{c|}{0.0047} & 47.95 & \multicolumn{1}{c|}{0.0047} & 50.73 & \multicolumn{1}{c|}{0.0038} & 44.56 \\ \hline
\end{tabular}
}
\caption{AUC and RIG performance of Two-steps Joint-training on two data sets} \label{table:jointtraining}
\end{table*}
Table \ref{table:jointtraining} shows the AUC/RIG performance of these three settings on both data sets. From this table, one can  find that  two-steps joint-training brings significant gain for both data sets. On Bing Ads data, the AUC gain is more than 0.3\% , and more than 0.4\% over KDD CUP 2012 data. All these gains are shown by the t-tests to be significant.

\subsubsection{Aggregated Training Loss}  \label{subsubsec:exp_trainingloss}
\begin{figure}[h]
  \vspace{-2 mm}
  \centering
  \includegraphics[width=\linewidth]{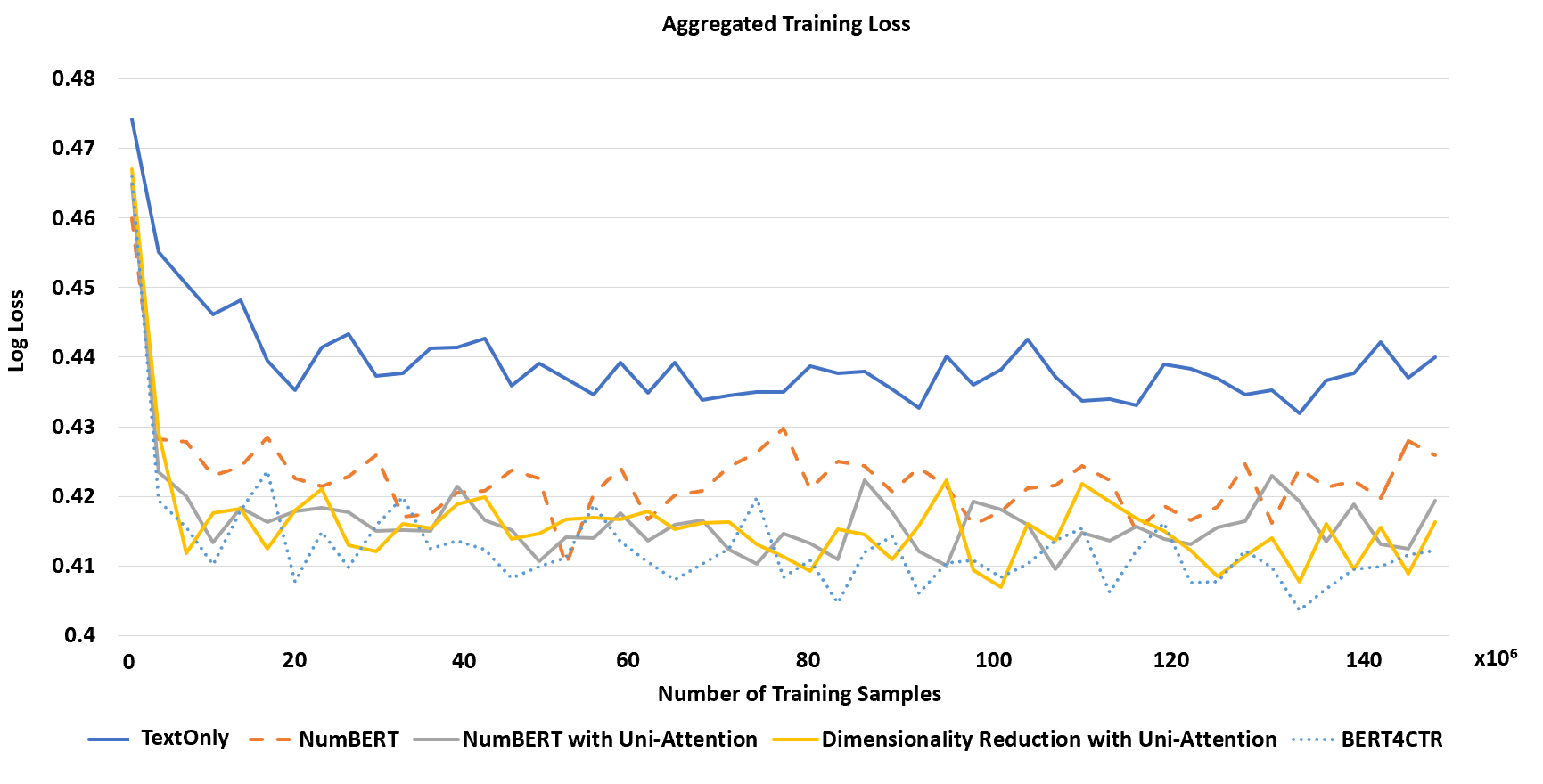}
  \caption{Curves of aggregated training loss  on Bing Ads data} \label{fig:training_loss}
  \vspace{-2 mm}
\end{figure}

We show in Figure \ref{fig:training_loss} the evaluation of training loss for all the alternative models. The aggregated log-loss is derived after training per million samples, and the trends are reported in Figure \ref{fig:training_loss} for the  first training epoch over Bing Ads data set.

The figure leads to four observations. First, the training loss of the model without non-textual features ({\em i.e.}, TextOnly model) is higher than the ones of the other alternative models, indicating that non-textual features are important for CTR prediction. Second, training loss for NumBERT with uni-attention is below the one of NumBERT, which provides another evidence that uni-attention architecture improves CTR prediction. Third, the training loss curves for NumBERT with uni-attention are close, with and without  dimensionality reduction. This means  that dimensionality reduction does not compromise the accuracy of prediction much while reducing the time-costs of training and inference. Finally, training loss for BERT4CTR is the lowest one , showing clearly that the two-steps joint-training improves the performance of CTR prediction. These observations are  consistent with the ones obtained  based on AUC and RIG metrics.

\subsection{Comparison of BERT4CTR with Other Multi-modal Frameworks}
In this part, we compare BERT4CTR performances with the three alternative frameworks, cascading framework, Shallow Interaction framework and NumBERT,  that can handle multi-modal inputs for CTR prediction.
\begin{table*}[!htbp]
\subtable[AUC Performance]{
\resizebox{2.1\columnwidth}{!}{
\begin{tabular}{|c|c|ccc|cc|c|c|cc|cc|cc|cc|cc|cc|}
\hline
\multirow{4}{*}{Dataset} & \multirow{4}{*}{Slice} & \multicolumn{3}{c|}{Cascading Framework} & \multicolumn{2}{c|}{Shallow Interaction Framework} & NumBERT Framework & BERT4CTR Framework & \multicolumn{2}{c|}{\multirow{2}{*}{$\Delta$ $AUC_{M7-M1}$}} & \multicolumn{2}{c|}{\multirow{2}{*}{$\Delta$ $AUC_{M7-M2}$}} & \multicolumn{2}{c|}{\multirow{2}{*}{$\Delta$ $AUC_{M7-M3}$}} & \multicolumn{2}{c|}{\multirow{2}{*}{$\Delta$ $AUC_{M7-M4}$}} & \multicolumn{2}{c|}{\multirow{2}{*}{$\Delta$ $AUC_{M7-M5}$}} & \multicolumn{2}{c|}{\multirow{2}{*}{$\Delta$ $AUC_{M7-M6}$}} \\ \cline{3-9}
 &  & \multicolumn{1}{c|}{Model 1} & \multicolumn{1}{c|}{Model 2} & Model 3 & \multicolumn{1}{c|}{Model 4} & Model 5 & Model 6 & Model 7 & \multicolumn{2}{c|}{} & \multicolumn{2}{c|}{} & \multicolumn{2}{c|}{} & \multicolumn{2}{c|}{} & \multicolumn{2}{c|}{} & \multicolumn{2}{c|}{} \\ \cline{3-9}
 &  & \multicolumn{1}{c|}{Wide \& Deep} & \multicolumn{1}{c|}{DeepFM} & NN + GBDT & \multicolumn{1}{c|}{\begin{tabular}[c]{@{}c@{}}Shallow Interaction\\ - 1 Layer\end{tabular}} & \begin{tabular}[c]{@{}c@{}}Shallow Interaction\\ - N Layers\end{tabular} & NumBERT & BERT4CTR  & \multicolumn{2}{c|}{} & \multicolumn{2}{c|}{} & \multicolumn{2}{c|}{} & \multicolumn{2}{c|}{} & \multicolumn{2}{c|}{} & \multicolumn{2}{c|}{} \\ \cline{3-21} 
 &  & \multicolumn{1}{c|}{AUC} & \multicolumn{1}{c|}{AUC} & AUC & \multicolumn{1}{c|}{AUC} & AUC & AUC & AUC & \multicolumn{1}{c|}{Diff} & T & \multicolumn{1}{c|}{Diff} & T & \multicolumn{1}{c|}{Diff} & T & \multicolumn{1}{c|}{Diff} & T & \multicolumn{1}{c|}{Diff} & T & \multicolumn{1}{c|}{Diff} & T \\ \hline
\multirow{2}{*}{Bing Ads} & ALL & \multicolumn{1}{c|}{0.8932} & \multicolumn{1}{c|}{0.8961} & \multicolumn{1}{c|}{0.8956} & \multicolumn{1}{c|}{0.8968} & 0.8976 & 0.8961 & 0.9014 & \multicolumn{1}{c|}{0.0082} & 73.42 & \multicolumn{1}{c|}{0.0053} & 52.64 & \multicolumn{1}{c|}{0.0058} & 55.37 & \multicolumn{1}{c|}{0.0046} & 53.47 & \multicolumn{1}{c|}{0.0038} & 47.15 & \multicolumn{1}{c|}{0.0053} & 60.37 \\ \cline{2-21} 
 & Tail & \multicolumn{1}{c|}{0.8043} & \multicolumn{1}{c|}{0.8078} & \multicolumn{1}{c|}{0.8070} & \multicolumn{1}{c|}{0.8084} & 0.8097 & 0.8078 & 0.8136 & \multicolumn{1}{c|}{0.0093} & 72.68 & \multicolumn{1}{c|}{0.0058} & 60.13 & \multicolumn{1}{c|}{0.0066} & 61.32 &  \multicolumn{1}{c|}{0.0052} & 52.76 & \multicolumn{1}{c|}{0.0039} & 40.89 & \multicolumn{1}{c|}{0.0058} & 67.51 \\ \hline
\multirow{2}{*}{KDD CUP} & ALL & \multicolumn{1}{c|}{0.8223} & \multicolumn{1}{c|}{0.8278} &  \multicolumn{1}{c|}{0.8269} & \multicolumn{1}{c|}{0.8286} & 0.8304 & 0.8273 & 0.8347 & \multicolumn{1}{c|}{0.0124} & 95.73 & \multicolumn{1}{c|}{0.0069} & 47.05 & \multicolumn{1}{c|}{0.0078} &   69.98 & \multicolumn{1}{c|}{0.0061} & 74.33 & \multicolumn{1}{c|}{0.0043} & 51.58 & \multicolumn{1}{c|}{0.0074} & 69.90 \\ \cline{2-21} 
 & Tail & \multicolumn{1}{c|}{0.7471} & \multicolumn{1}{c|}{0.7531} & \multicolumn{1}{c|}{0.7526} & \multicolumn{1}{c|}{0.7537} & 0.7567 & 0.7521 & 0.7618 & \multicolumn{1}{c|}{0.0147} & 106.42 & \multicolumn{1}{c|}{0.0087} & 65.09 & \multicolumn{1}{c|}{0.0092} & 71.48 & \multicolumn{1}{c|}{0.0081} & 74.68 & \multicolumn{1}{c|}{0.0051} & 60.46 & \multicolumn{1}{c|}{0.0097} & 79.34 \\ \hline
\end{tabular}
}
}
\subtable[RIG Performance]{
\resizebox{2.1\columnwidth}{!}{
\begin{tabular}{|c|c|ccc|cc|c|c|cc|cc|cc|cc|cc|cc|}
\hline
\multirow{4}{*}{Dataset} & \multirow{4}{*}{Slice} & \multicolumn{3}{c|}{Cascading Framework} & \multicolumn{2}{c|}{Shallow Interaction Framework} & NumBERT Framework & BERT4CTR Framework & \multicolumn{2}{c|}{\multirow{2}{*}{$\Delta$ $RIG_{M7-M1}$}} & \multicolumn{2}{c|}{\multirow{2}{*}{$\Delta$ $RIG_{M7-M2}$}} & \multicolumn{2}{c|}{\multirow{2}{*}{$\Delta$ $RIG_{M7-M3}$}} & \multicolumn{2}{c|}{\multirow{2}{*}{$\Delta$ $RIG_{M7-M4}$}} & \multicolumn{2}{c|}{\multirow{2}{*}{$\Delta$ $RIG_{M7-M5}$}} & \multicolumn{2}{c|}{\multirow{2}{*}{$\Delta$ $RIG_{M7-M6}$}} \\ \cline{3-9}
 &  & \multicolumn{1}{c|}{Model 1} & \multicolumn{1}{c|}{Model 2} & Model 3 & \multicolumn{1}{c|}{Model 4} & Model 5 & Model 6 & Model 7 & \multicolumn{2}{c|}{} & \multicolumn{2}{c|}{} & \multicolumn{2}{c|}{} & \multicolumn{2}{c|}{} & \multicolumn{2}{c|}{} & \multicolumn{2}{c|}{} \\ \cline{3-9}
 &  & \multicolumn{1}{c|}{Wide \& Deep} & \multicolumn{1}{c|}{DeepFM} & NN + GBDT & \multicolumn{1}{c|}{\begin{tabular}[c]{@{}c@{}}Shallow Interaction\\ - 1 Layer\end{tabular}} & \begin{tabular}[c]{@{}c@{}}Shallow Interaction\\ - N Layers\end{tabular} & NumBERT & BERT4CTR  & \multicolumn{2}{c|}{} & \multicolumn{2}{c|}{} & \multicolumn{2}{c|}{} & \multicolumn{2}{c|}{} & \multicolumn{2}{c|}{} & \multicolumn{2}{c|}{} \\ \cline{3-21} 
 &  & \multicolumn{1}{c|}{RIG} & \multicolumn{1}{c|}{RIG} & RIG & \multicolumn{1}{c|}{RIG} & RIG & RIG & RIG & \multicolumn{1}{c|}{Diff} & T & \multicolumn{1}{c|}{Diff} & T & \multicolumn{1}{c|}{Diff} & T & \multicolumn{1}{c|}{Diff} & T & \multicolumn{1}{c|}{Diff} & T & \multicolumn{1}{c|}{Diff} & T \\ \hline
\multirow{2}{*}{Bing Ads} & ALL & \multicolumn{1}{c|}{0.5321} & \multicolumn{1}{c|}{0.5356} & \multicolumn{1}{c|}{0.5349} & \multicolumn{1}{c|}{0.5360} & 0.5372 & 0.5348 & 0.5413 & \multicolumn{1}{c|}{0.0092} & 78.46 & \multicolumn{1}{c|}{0.0057} & 54.89 & \multicolumn{1}{c|}{0.0064} & 61.73 & \multicolumn{1}{c|}{0.0053} & 55.82 & \multicolumn{1}{c|}{0.0041} & 51.34 & \multicolumn{1}{c|}{0.0065} & 68.89 \\ \cline{2-21} 
 & Tail & \multicolumn{1}{c|}{0.4702} & \multicolumn{1}{c|}{0.4734} & \multicolumn{1}{c|}{0.4723} & \multicolumn{1}{c|}{0.4726} & 0.4738 & 0.4719 & 0.4801 & \multicolumn{1}{c|}{0.0099} & 80.43 & \multicolumn{1}{c|}{0.0067} & 57.67 & \multicolumn{1}{c|}{0.0078} & 68.34 & \multicolumn{1}{c|}{0.0075} & 65.19 & \multicolumn{1}{c|}{0.0063} & 50.38 & \multicolumn{1}{c|}{0.0082} & 75.33 \\ \hline
\multirow{2}{*}{KDD CUP} & ALL & \multicolumn{1}{c|}{0.4794} & \multicolumn{1}{c|}{0.4829} & \multicolumn{1}{c|}{0.4822} & \multicolumn{1}{c|}{0.4842} & 0.4853 & 0.4827 & 0.4903 & \multicolumn{1}{c|}{0.0109} & 100.25 & \multicolumn{1}{c|}{0.0074} & 71.26 & \multicolumn{1}{c|}{0.0081} & 80.16 & \multicolumn{1}{c|}{0.0061} & 75.82 & \multicolumn{1}{c|}{0.0050} & 56.63 & \multicolumn{1}{c|}{0.0076} & 67.79 \\ \cline{2-21} 
 & Tail & \multicolumn{1}{c|}{0.3689} & \multicolumn{1}{c|}{0.3726} & \multicolumn{1}{c|}{0.3719} & \multicolumn{1}{c|}{0.3739} & 0.3754 & 0.3724 & 0.3821 & \multicolumn{1}{c|}{0.0132} & 107.94 & \multicolumn{1}{c|}{0.0095} & 91.30 & \multicolumn{1}{c|}{0.0102} & 96.59 & \multicolumn{1}{c|}{0.0082} & 76.54 & \multicolumn{1}{c|}{0.0067} & 61.09 & \multicolumn{1}{c|}{0.0097} & 81.33 \\ \hline
\end{tabular}
}
}
\caption{AUC and RIG performance of BERT4CTR compared with representative models in use on two data sets} \label{table:auc_summary}
\end{table*}
In Table \ref{table:auc_summary}, the  AUC and RIG performance are shown for all possible alternatives. Three major observations can be extracted from this table. First, cross-information learning between textual and non-textual features during fine-tuning phase can improve the accuracy of prediction significantly, {\em e.g.}, BERT4CTR brings more than 0.5\% AUC gain on both Bing Ads data and KDD CUP 2012 data, compared with all cascading methods (Wide \& Deep, DeepFM and NN+GBDT). Second, although increasing the depth of network in non-textual part improves the accuracy of CTR prediction, deep uni-attentions between textual features and non-textual features still bring considerable  improvement for CTR prediction, {\em e.g.}, BERT4CTR can bring 0.4\% AUC gain on both Bing Ads data and KDD CUP 2012 data, compared with Shallow Interaction-N Layers model, showing the pure benefits brought by the uni-attention architecture. Finally, among all seven alternative models in Table \ref{table:auc_summary}, BERT4CTR shows the highest AUC on both data sets, which gives evidence that the design presented in Section \ref{sec:bert-ctr} is an effective way to learn the cross-information between multi-modal inputs for CTR prediction.

\begin{table*}[!htbp]
\subtable[Training Cost]{ \label{subtable:latency_summary_training}
\resizebox{1.8\columnwidth}{!}{
\begin{tabular}{|c|c|cc|cc|cc|cc|}
\hline
\multirow{2}{*}{Framework} & \multirow{2}{*}{Model} & \multicolumn{2}{c|}{Average} & \multicolumn{2}{c|}{Median} & \multicolumn{2}{c|}{90th Percentile} & \multicolumn{2}{c|}{95th Percentile}  \\ \cline{3-10} 
&  & \multicolumn{1}{c|}{Bing Ads} & KDD CUP & \multicolumn{1}{c|}{Bing Ads} & KDD CUP & \multicolumn{1}{c|}{Bing Ads} & KDD CUP & \multicolumn{1}{c|}{Bing Ads} & KDD CUP \\ \hline
\multicolumn{1}{|c|}{\multirow{2}{*}{\begin{tabular}[c]{@{}c@{}}Shallow Interaction \\Framework (Two-steps)\end{tabular}}} & \begin{tabular}[c]{@{}c@{}}Shallow Interaction\\ - 1 Layer\end{tabular} & \multicolumn{1}{c|}{19.72} & 7.69 & \multicolumn{1}{c|}{19.66} & 7.67 & \multicolumn{1}{c|}{19.76} & 7.72 & \multicolumn{1}{c|}{19.87} & 7.75  \\ \cline{2-10}
\multicolumn{1}{|c|}{} & \begin{tabular}[c]{@{}c@{}}Shallow Interaction\\ - N Layers\end{tabular} & \multicolumn{1}{c|}{20.54} & 8.24 & \multicolumn{1}{c|}{20.42} & 8.22 & \multicolumn{1}{c|}{20.66} & 8.26 & \multicolumn{1}{c|}{20.75} & 8.33  \\ \hline
NumBERT Framework & NumBERT & \multicolumn{1}{c|}{60.50} & 12.36 & \multicolumn{1}{c|}{60.11} & 12.29 & \multicolumn{1}{c|}{60.61} & 12.45 & \multicolumn{1}{c|}{60.89} & 12.58  \\ \hline
BERT4CTR Framework (Two-steps) & BERT4CTR  & \multicolumn{1}{c|}{22.06} & 8.69 & \multicolumn{1}{c|}{21.88} & 8.64 & \multicolumn{1}{c|}{22.32} & 8.77 & \multicolumn{1}{c|}{22.45} & 8.85 \\ \hline
\end{tabular}
}
}
\subtable[Inference Cost] { \label{subtable:latency_summary_validation}
\resizebox{1.8\columnwidth}{!}{
\begin{tabular}{|c|c|cc|cc|cc|cc|}
\hline
\multirow{2}{*}{Framework} & \multirow{2}{*}{Model} & \multicolumn{2}{c|}{Average} & \multicolumn{2}{c|}{Median} & \multicolumn{2}{c|}{90th Percentile} & \multicolumn{2}{c|}{95th Percentile}  \\ \cline{3-10} 
& & \multicolumn{1}{c|}{Bing Ads} & KDD CUP & \multicolumn{1}{c|}{Bing Ads} & KDD CUP & \multicolumn{1}{c|}{Bing Ads} & KDD CUP & \multicolumn{1}{c|}{Bing Ads} & KDD CUP \\ \hline
\multicolumn{1}{|c|}{\multirow{3}{*}{Cascading Framework}} & Wide \& Deep & \multicolumn{1}{c|}{ 6.09 } & 2.79  & \multicolumn{1}{c|}{ 6.06 } &  2.76 & \multicolumn{1}{c|}{ 6.19 } & 2.88 & \multicolumn{1}{c|}{6.30 } &  2.98 \\ \cline{2-10} 
\multicolumn{1}{|c|}{} & DeepFM & \multicolumn{1}{c|}{ 6.18 } & 2.81  & \multicolumn{1}{c|}{ 6.14 } &  2.78 & \multicolumn{1}{c|}{ 6.27 } & 2.91  & \multicolumn{1}{c|}{ 6.39 } &  3.00  \\ \cline{2-10} 
\multicolumn{1}{|c|}{} & NN+GBDT & \multicolumn{1}{c|}{ 6.14 } & 2.81  & \multicolumn{1}{c|}{ 6.12 } & 2.77  & \multicolumn{1}{c|}{ 6.23 } & 2.90  & \multicolumn{1}{c|}{ 6.35 } &  3.01  \\ \hline
\multicolumn{1}{|c|}{\multirow{2}{*}{Shallow Interaction Framework}} & \begin{tabular}[c]{@{}c@{}}Shallow Interaction\\ - 1 Layer\end{tabular} & \multicolumn{1}{c|}{6.05} & 2.82 & \multicolumn{1}{c|}{6.01} & 2.78 & \multicolumn{1}{c|}{6.13} & 2.87 & \multicolumn{1}{c|}{6.29} & 2.96  \\ \cline{2-10} 
\multicolumn{1}{|c|}{} & \begin{tabular}[c]{@{}c@{}}Shallow Interaction\\ - N Layers\end{tabular} & \multicolumn{1}{c|}{6.57} & 3.08 & \multicolumn{1}{c|}{6.52} & 3.04 & \multicolumn{1}{c|}{6.66} & 3.13 & \multicolumn{1}{c|}{6.75} & 3.22  \\ \hline
NumBERT Framework & \multicolumn{1}{|c|}{NumBERT} & \multicolumn{1}{c|}{14.73} & 4.72 & \multicolumn{1}{c|}{14.70} & 4.68 & \multicolumn{1}{c|}{14.83} & 4.81 & \multicolumn{1}{c|}{14.88} & 4.92  \\ \hline
BERT4CTR Framework & \multicolumn{1}{|c|}{BERT4CTR} & \multicolumn{1}{c|}{7.05} & 3.36 & \multicolumn{1}{c|}{7.03} & 3.32 & \multicolumn{1}{c|}{7.19} & 3.42 & \multicolumn{1}{c|}{7.31} & 3.51  \\ \hline
\end{tabular}
}
}
\caption{Time-cost performance (ms/sample) of BERT4CTR compared with representative models in use on two data sets} \label{table:latency_summary}
\end{table*}

The time-costs of training and inference for the alternative models are shown in Table \ref{table:latency_summary}.
At first, we observe that BERT4CTR does not bring  significant increases in training time compared with Shallow Interaction. In detail, the training time of BERT4CTR only increases by 7\%  compared with Shallow Interaction-N Layers and by 14\% compared with Shallow Interaction-1 Layer that have been widely used in industry \cite{wang2022learning}\cite{chen2019behavior}. For example, in Microsoft Bing Ads, Shallow Interaction-1 Layer framework is used  to refresh a RoBERTa-24 model. The training takes 5 days in one cycle. According to Table \ref{table:latency_summary}, BERT4CTR will take 5.8 days on the same settings, that is still less than the weekly re-calibration deadline.

The inference delay of BERT4CTR is close to cascading, and Shallow Interaction framework, and much less than NumBERT, {\em e.g.}, BERT4CTR can reduce inference delay by  52\% (resp., 29\%) on Bing Ads data (resp., KDD CUP 2012 data), compared with NumBERT.  

The results from Table \ref{table:auc_summary} and Table \ref{table:latency_summary} give strong evidences that BERT4CTR can achieve both high accuracy and low training and inference delay for CTR prediction.




\section{Discussion}
Although, we used in this paper the CTR prediction with numerical features as our main applicative scenario, the framework of BERT4CTR proposed is applicable to other scenarios mixing textual and non-textual features. For example, one can extract the representative embedding from images through VGGNet \cite{simonyan2014very}, ResNet \cite{he2016deep} {\em etc.} to replace the token $E_E$ in Figure \ref{fig:joint-training} and calculate the uni-attentions. 
Besides, the Knowledge-Distillation technology \cite{hinton2015distilling} can be adopted on BERT4CTR, where a light model handling textual and non-textual inputs, with uni-attention and dimensionality reduction, can be learned under the supervision of predicted scores from a well-trained BERT4CTR model with deep layers.

\section{Conclusion} \label{sec:conclusion}
In this paper, we focused on the design of an efficient framework to combine pre-trained language model with non-textual features for CTR prediction. We started from NumBERT which is the traditional usage
of pre-trained language model to integrate textual features and numerical ones, and introduced three improvements, uni-attention, dimensionality reduction and two-steps joint-training, to compose a novel framework BERT4CTR. Comprehensive experiments on both commercial data and public data showed that BERT4CTR can achieve significant gains in accuracy of prediction while keeping low time-costs of training and inference, and therefore provide a promising solution of CTR prediction in the real world.


\bibliographystyle{ACM-Reference-Format}
\bibliography{acmart}

\appendix 

\section{appendix}
For the reproducibility, we provide the important details about experiments on the public KDD CUP 2012 data set, including the data description, 
 the settings on textual and non-textual part, and the pseudo code of uni-attention.

\subsection{Data Details} \label{subsec:data details}
The data set contains 235 million search ads impressions sampled from session logs of Tencent search engine Soso.com. Each sample in this data set is likewise Bing Ads data and contains five components: 
\begin{itemize}
\item Query text: which is a list of anonymous tokens hashed from natural language; 
\item Ad text: which includes ad title and ad display URL and are also lists of anonymous tokens; 
\item CTR prediction features: there are 56 CTR prediction features including sparse features such as UserID, AdID, user's gender {\em etc.}, and dense features such as historical CTRs,  number of impressions per Ad/Query/User {\em etc.}; 
\item Position feature: a special feature which indicates the impressed position of this ad; 
\item Click label: where 1 means this ad has been clicked and 0 is not clicked. 
\end{itemize}

The 56 non-textual CTR features used in BERT4CTR are generated from the following ways:
\begin{itemize}
\item ID features: we mapped each raw ID attribute (such as AdID, QueryID, UserID, Gender, Age {\em etc.}) appeared in training data to a consecutive number. To consider the applicability of model on validation data where some new IDs are not appeared in training data, we used the robust training, {\em i.e.}, we randomly removed 5\% IDs in training data and set them as "Missing", which is a special mapped number. In inference, we also set the new IDs in validation data as "Missing" to make the model work well; 

\item Historical features: we calculated the historical CTRs and number of impressions from different perspectives, including the Ad-level, User-level, Query-level, Gender-level and Age-level {\em etc.};

\item Length features: these length features include Query Length, AdTitle Length, AdDesription Length and Keyword Length {\em etc.};

\item Semantic features: we calculated  TF $\times$ IDF value for each token (such as Query tokens, AdTitle tokens, AdDescription tokens {\em etc.}) provided in training data, as a type of semantic feature.

\end{itemize}

As there is no time information in this data set, it is impossible to split the training data and validation data according to impression time. Alternatively, we take a simple way to generate the training and validation set, where we randomly choose 1/11 of samples as validation data and the remaining as training data. Table \ref{table:table8} summaries the statistics of training data and validation data.

\begin{table}[h]
\begin{tabular}{|c|c|c|c|}
\hline
 & Impressions & Clicks & CTR \\ \hline
Training Data & 216,038,149 & 7,550,609 & 0.0349 \\ \hline
Validation Data & 19,544,730 & 667,024 & 0.0341 \\ \hline
\end{tabular}
\caption{Statistics for KDD CUP 2012 data set}
\label{table:table8}
\end{table}

\subsection{Settings on Textual Part}
\label{subsec:textual_settings}
We choose BERT-12 model as initial model where query and ad title concatenated with ad display URL are used as two input sentences. As the privacy, each textual token in KDD CUP 2012 data set is anonymous to one hash ID  and therefore, it is difficult to map the hash ID from data to vocabulary ID of pre-trained language models. To solve this issue, we treat these anonymous tokens as new words, where we calculate the TF $\times$ IDF for each anonymous token and choose the top 300 thousands of these to build a new vocabulary. The masking rate is 15\% for MLM task and Adam optimizer with learning rate of $1 \times 10^{ 
-4}$ is used in both pre-training and fine-tuning phase. 
After two epochs of pre-training and four epochs of fine-tuning with this new vocabulary, the AUC of BERT model can achieve 75.91\% on ALL slice, meaning that our solution is effective.

\subsection{Settings on Non-textual Part} \label{subsec:nontextual_settings}
As described in Section \ref{sec:bert-ctr}, 
the processing methods for sparse features and dense ones are different.  For sparse features, we extent each input to a 32-dimensional embedding respectively. While for dense features, we normalize them with max-min normalization at first and then expand each normalized value into 101-dimensional one-hot vectors based on 0.01 buckets. Afterwards the 32-dimensional embedding is generated by looking up embedding table. Totally a 1792-dimensional embedding is generated by concatenating all of 56 32-dimensional sub-embeddings (excluding the embedding from position feature) and then a 512-dimensional hidden layer is followed as the final embedding layer to calculate uni-attention.

\begin{figure}[h]
  \vspace{-1 mm}
  \centering
  \includegraphics[width=\linewidth]{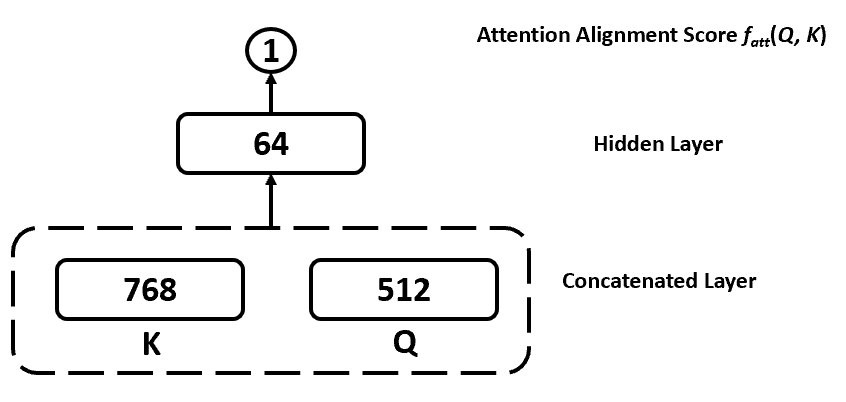}
  \caption{The network to calculate the attention alignment score between Query and Key in uni-attention} \label{fig:additive_attention}
  \vspace{-2 mm}
\end{figure}

For uni-attention, we set the dimension of hidden layer in Equation \ref{eq:additive_attention}, denoted as $d_a$, as 64, and therefore the network to calculate the attention alignment score between Query and Key in uni-attention (based on BERT-12) is depicted in Figure \ref{fig:additive_attention}.

\label{subsec:nontextual_settings}

\subsection{Implementation of Uni-Attention}

We also provide the details on the implementation of uni-attention here, which is the core of BERT4CTR. In one layer of BERT4CTR, the output from the non-textual token after dimensionality reduction can be denoted as $X \in \mathbb{R}^{d_X \times 1}$, where $d_X$ is the dimension of output for this non-textual token (512 in our experiment). Beside, the outputs from textual tokens can be denoted as $Y \in \mathbb{R}^{d_Y \times l_Y}$, where $d_Y$ is the dimension of output for each textual token (768 in our experiment) and $l_Y$ is the maximum sequence length of textual input (64 in our experiment). We also use the same Attention-Mask mechanism as the one used in textual part, for the calculation of uni-attention, where $Mask \in [0, 1]^{l_Y}$. 

The pseudo code of uni-attention based on additive attention is shown in Algorithm \ref{algo:code}, in which $W_q$, $b_q$, $W_k$, $b_k$, $W_v$, $b_v$, $W_a$, $b_a$, $W_i$ and $b_i$ are trainable parameters.

\floatname{algorithm}{Algorithm}
\renewcommand{\algorithmicrequire}{\textbf{Input:}}
\renewcommand{\algorithmicensure}{\textbf{Output:}}

    \begin{algorithm}[H]
        \caption{$U \gets UniAttention (X, Y, Mask)$} \label{algo:code}
        \begin{algorithmic}[1] 
            \Require $X \in \mathbb{R}^{d_X \times 1}$, $Y \in \mathbb{R}^{d_Y \times l_Y}$, $Mask \in [0, 1]^{l_Y}$
            \Ensure $U \in  \mathbb{R}^{d_X \times 1}$ 

            \Function {UniAttention}{$X$, $Y$, $Mask$}
                \State $Q \gets W_qX+b_q\mathbbm{1}^T$ \Comment{$W_q \in \mathbb{R}^{d_X \times d_X}$, $b_q \in \mathbb{R}^{d_X}$}
                \State $K \gets W_kY+b_k\mathbbm{1}^T$ \Comment{$W_k \in \mathbb{R}^{d_Y \times d_Y}$, $b_k \in \mathbb{R}^{d_Y}$}
                \State $V \gets W_vY+b_v\mathbbm{1}^T$ \Comment{$W_v \in \mathbb{R}^{d_X \times d_Y}$, $b_v \in \mathbb{R}^{d_X}$}
                \State $Repeat(Q, l_Y)$ \Comment{$Q \in \mathbb{R}^{d_X \times l_Y}$}
                \State $M \gets Concat(Q, K)$ \Comment{$M \in \mathbb{R}^{(d_X+d_Y) \times l_Y}$}
                \State $H \gets Tanh(W_aM+b_a\mathbbm{1}^T)$ \Comment{$H \in \mathbb{R}^{d_a \times l_Y}$}
                \State $S \gets W_iH+b_i\mathbbm{1}^T$ \Comment{$S \in \mathbb{R}^{1 \times l_Y}$}
                \State $\forall S_i \in S, if~\neg Mask[i],~then~S_i \gets -\infty$ 
                \State $U \gets V~Softmax(S^T)$ \Comment{$U \in \mathbb{R}^{d_X \times 1}$}
                \State \Return $U$ 
            \EndFunction
           
        \end{algorithmic}
    \end{algorithm}

\end{document}